%% file: main.tex
\documentclass{article}
\usepackage{amsmath}
\usepackage{makecell}

\usepackage[preprint    ]{neurips_2024}

\usepackage[utf8]{inputenc} 
\usepackage[T1]{fontenc}    
\usepackage{hyperref}       
\usepackage{url}            
\usepackage{booktabs}       
\usepackage{amsfonts}       
\usepackage{nicefrac}       
\usepackage{microtype}      
\usepackage{xcolor}         
\usepackage{graphicx}
\usepackage{import}
\usepackage{tabularx}
\usepackage{xspace} 
\usepackage[textwidth=1.7cm,textsize=small]{todonotes}

\input{definitions}

\def\secref#1{Section~\ref{#1}}
\def\tabref#1{Table~\ref{#1}}

\title{Improving generalisability of 3D binding affinity models in low data regimes}

\author{%
  Julia Buhmann\thanks{Equal contribution.} \\
  Exscientia\\
  \texttt{jbuhmann@exscientia.co.uk} \\
  \And
  Ward Haddadin\footnotemark[1] \\
  Exscientia\\
  \texttt{whaddadin@exscientia.co.uk} \\
  \And
  Luk\'a\v{s} Pravda \\
  Exscientia\\
  \texttt{lpravda@exscientia.co.uk} \\
    \And
  Alan Bilsland \\
  Exscientia\\
  \texttt{abilsland@exscientia.co.uk} \\
    \And
  Hagen Triendl\\
  Exscientia\\
  \texttt{ hagentriendl@gmail.com} \\
}

\begin{document}

\maketitle




\import{sections/}{0_abstract}
\import{sections/}{1_introduction}
\import{sections/}{2_methods}
\import{sections/}{3_results}
\import{sections/}{4_discussion}

\import{sections/}{5_acknowledgements}

\bibliography{references}
\bibliographystyle{plainnat}


\newpage

\appendix

\import{sections/}{supplementary_info}


\clearpage


\end{document}

%% file: definitions.tex
\def\rfscore{RF-Score\xspace}
\def\onion{OnionNet-2\xspace}
\def\singleprotein{Single-Protein\xspace}
\def\ligandbias{Ligand-Bias\xspace}
\def\mw{Molecular-Weight\xspace}

\def\singlegraph{Single-Graph\xspace}
\def\multigraph{Multi-Graph\xspace}

\def\pdbbind{PDBBind\xspace}
\def\otherproteins{Other-Proteins\xspace}
\def\casestudy{Case-Study-Proteins\xspace}

\def\oursplit{Low-Sim\xspace}

\def\physicsml{\textsc{physicsml}\xspace}
\def\molflux{\textsc{molflux}\xspace}

\def\lowsimgithub{\url{https://github.com/Exscientia/low-sim-pdbbind}\xspace}
\def\ccdczenodo{\url{https://zenodo.org/records/13772124}}

\makeatletter
  \if@submission
    \def\lowsimgithub{URL anonymized\xspace}
    \def\ccdczenodo{URL anonymized\xspace}
    \def\physicsml{anonymized\xspace}
    \def\molflux{anonymized\xspace}
  \fi
\makeatother

\makeatletter  
  \if@submission  
    \newcommand{\anonref}[1]{\cite{anon}}  
  \else  
    \newcommand{\anonref}[1]{\cite{#1}}  
  \fi  
\makeatother

%% file: sections/0_abstract.tex
\begin{abstract}
    Predicting protein-ligand binding affinity is an essential part of computer-aided drug design. However, generalisable and performant global binding affinity models remain elusive, particularly in low data regimes. Despite the evolution of model architectures, current benchmarks are not well-suited to probe the generalisability of 3D binding affinity models. Furthermore, 3D global architectures such as GNNs have not lived up to performance expectations. To investigate these issues, we introduce a novel split of the \pdbbind dataset, minimizing similarity leakage between train and test sets and allowing for a fair and direct comparison between various model architectures.
    On this low similarity split, we demonstrate that, in general, 3D global models are superior to protein-specific local models in low data regimes. We also demonstrate that the performance of GNNs benefits from three novel contributions: supervised pre-training via quantum mechanical data, unsupervised pre-training via small molecule diffusion, and explicitly modeling hydrogen atoms in the input graph. We believe that this work introduces promising new approaches to unlock the potential of GNN architectures for binding affinity modelling. 
\end{abstract}

%% file: sections/1_introduction.tex
\section{Introduction}
Computer-aided drug design relies on the accurate prediction of protein-ligand binding affinity to achieve a therapeutic effect, ensuring selectivity against other proteins and avoiding off-target toxicity \citep{kairys2019binding}.

Apart from classical approaches for binding affinity prediction (usually docking methods which use a combination of empirical molecular and statistical force fields \citep{Jones1997Gold, Eberhardt2021Vina}), a diverse array of machine learning (ML) strategies have been proposed in the last decade. 
There has been increasing interest in developing ML models that use 3D data of protein-ligand complexes as input. In principle, these models are the best suited for predicting binding affinity since they should be able to capture fundamental interaction mechanisms such as hydrogen bonds or hydrophobic and ionic interactions between the protein and the ligand.
Numerous types of 3D ML models are presented in the literature. In \cite{ballester2010rfscore} and \cite{Wang2021onionnet2}, the protein and ligand interactions were condensed into contact map features and used in tree-based models (Random Forest) and convolutional neural networks (CNN). In \cite{volkov2022frustration}, they use a graph neural network (GNN) which uses the protein-ligand graph as input. They also propose including interacting nodes in the graph to indicate known interactions between protein and ligand atoms. Using GNNs as well, \cite{Zhang2023SSGNN} add thresholds to the distance encoding to avoid overfitting on small distance variations.

The majority of the literature indicates that it is not yet clear whether a specific model type consistently achieves the best results \citep{durant2023robustly}. Part of this ambiguity is due to the lack of consistent benchmarks to evaluate the performance of the diverse array of models. The common dataset used for benchmarking 3D binding affinity models is the \pdbbind \citep{liu2015pdbbind}, a dataset of crystal structures from the Protein Data Bank (PDB) with curated binding affinity measurements. Over the last few years, many splits have been proposed for the \pdbbind dataset to probe performance and generalisation of different model types \citep{volkov2022frustration,durant2023robustly,li2024leakproofpdbbindreorganized}, but each come with their own drawbacks. 

Overall, the results indicate that simple models and baselines perform just as well as the more complicated 3D models that use structural information \citep{durant2023robustly}. This indicates that 3D models are not learning generalizable information but only dataset biases, hence they have not yet met their projected expectations \citep{volkov2022frustration}.

In this work we investigate the performance of binding affinity model families in a robust setting to probe what they learn and how they generalise. We compare 3D global models to protein-specific local models commonly used in real world drug discovery and also to baseline bias models. To achieve this, we propose a new split of the \pdbbind dataset based on protein and ligand similarity and constructed to suit bechmarking the various model families fairly and consistently. 
We use the new split and strong baselines to test multiple novel improvements to a plain 3D GNN model to push the boundaries of binding affinity modelling using 3D GNNs. 

We find that in low data regimes, 3D models significantly outperform protein-specific local models. With more data for a specific protein, local models quickly catch up. We also investigate the effect of hydrogen atoms on generalisability. As the structures in the \pdbbind are not consistently prepared, we use protein preparation software to prepare them consistently and include hydrogen atoms explicitly in the GNN encoding. We again find that at low data regimes, including hydrogen atoms explicitly is very important for generalisation. This advantage goes away with more data. Finally, we propose two pre-training methods to improve global 3D model performance. We pre-train GNNs on supervised quantum mechanical energy prediction and unsupervised small molecule diffusion. We show that both result in improvements at low data regimes. As far as we are aware, this is the first application of quantum mechanical pre-training and diffusion pre-training for binding affinity prediction.

%% file: sections/2_methods.tex
\section{Methods}

In this section, we present an overview of our methods and benchmarks. We discuss the dataset, structure preparation, model families benchmarked, and the new proposed split. We have made the code, prepared structures, and splits in this work publicly available at \lowsimgithub and \ccdczenodo.

\subsection{\pdbbind dataset}

We choose the \pdbbind dataset (release v2020\footnote{As of 2024, there is a newer \pdbbind release (v2021) available at \url{https://www.pdbbind-plus.org.cn/}. We did not use it in this work, as access is restricted.}) as our benchmarking data set \citep{liu2015pdbbind}. The data consists of crystal structures of bound protein-ligand complexes deposited in the PDB with curated binding affinity values ($\text{K}_\text{I}$, $\text{K}_\text{D}$ or $\text{IC}_{50}$). We use the protein-ligand subset of the general set from \pdbbind consisting of 19443 unique protein-ligand structures. The dataset is unbalanced. Many proteins have binding affinities measured only against a single ligand (one structure), some have measurements for a few ligands (few structures), and very few have measurements against more than 100 ligands.

It has been demonstrated, in both predictive and generative settings \citep{durant2023robustly,li2024leakproofpdbbindreorganized,buttenschoen2023posebustersaibaseddockingmethods}, that the splits routinely used in model benchmarking on \pdbbind contain data leakage. The similarity between proteins and ligands across training and test sets inflates metrics for certain models or tasks and makes rigorously probing their performance and generalisability difficult.

\subsection{Structure preparation} \label{dataset preparation}

We queried the PDB \citep{wwpdb2019protein} for biological assemblies of all the structures listed in the \pdbbind dataset.  Due to the nature of structure-determining techniques, the structures contain not only ligands of biological nature, but also residues of crystalization buffers or cryoprotectants. Hydrogen atoms are also mostly missing. 

To address both of these issues, each structure was prepared using CCDC software \citep{groom2016cambridge}, namely the Python API (\texttt{v.3.0.16}). Hydrogen atoms were added and the ligands listed by \pdbbind as biologically relevant were extracted into separate files. The remaining ligands of non-biological nature along with water molecules were removed.

After preparation, 18310 structures out of the total 19443 remained (1133 failed).

\subsection{Models}

Next, we discuss the models used in this study. An overview is presented in \tabref{table::models}. A full description of the hyperparameters of the features and models is provided in the Supplementary Information.

\begin{table}[h]  
\centering  
\caption{Overview of the different models used in this study. We indicate whether the models are global or local and whether they use 3D information or not. We make use of two baseline models to estimate dataset biases (\ligandbias and \mw).}  
\begin{tabular}{lccc}  
\toprule  
\textbf{Model Name} & \textbf{Global/Local} & \textbf{3D/non-3D} & \textbf{Input} \\  
\midrule  
\singleprotein & local & non-3D & ligand only \\ 

EGNN \citep{satorras2022enequivariantgraphneuralegnn}& global & 3D & pocket + ligand \\  
\rfscore \citep{ballester2010rfscore} & global & 3D & pocket + ligand \\  
\onion \citep{Wang2021onionnet2} & global & 3D & pocket + ligand    \\
\cmidrule(lr){1-4}
\ligandbias & global & non-3D & ligand only \\
\mw & local & non-3D & ligand  only \\
\bottomrule  
\end{tabular}
\label{table::models}
\end{table}

\subsubsection{Model families}

Before presenting the models used, we establish clear groups to classify them. There are numerous ways to categorize binding affinity model types, but in this work we introduce a particular grouping focused on two key aspects: the scope of application, global vs. local, and the type of data input, non-3D vs. 3D.

We denote models intended for use on different protein targets by \textbf{global models} and ones intended for use against a single protein target by \textbf{local models}. Local models are typically trained on ligand activity data measured against a single protein.
Binding affinity models can be further grouped according to how ligands and proteins are represented.
We denote models which do not use 3D coordinates by \textbf{non-3D models}. These models often use ligand descriptors such as fingerprints to encode ligands. If protein information is used, it is most commonly encoded with its amino acid sequence. \textbf{3D models} use the 3D coordinates of a ligand or a protein-ligand complex alongside non-3D information.

\subsubsection{\singleprotein local models}
\label{sec:methods:single_protein_local_models}

To compare to standard practices in the drug discovery community and in real world projects \citep{lo2018machine}, we benchmark the performance of \singleprotein local models built from ligand-based features and trained on binding activity data measured against a single protein. Analogous to established workflows \citep{jiang2021could,deng2023systematic}, we use classical ML models (Random Forest \citep{breiman2001randomforests}, XGBoost \citep{chen2016xgboost}, CatBoost \citep{prokhorenkova2018catboost}, Support Vector Machine \citep{cortes1995support}) combined with a selection of fingerprints (ECFP, FCFP, atom pairs, topological torsion) and molecular descriptors for features. More details on features and models are listed in the Supplementary Information.

\subsubsection{EGNN models}

We use the EGNN \citep{satorras2022enequivariantgraphneuralegnn} model\footnote{Although not included in this work, we also benchmarked more advanced 3D models such as MACE (\cite{batatia2023macehigherorderequivariantMACE}), Allegro (\cite{musaelian2022learninglocalequivariantrepresentationsAllegro}), and NequIP \cite{Batzner_2022Nequip}). However, we saw very poor performance. These models are top performers for large high quality datasets like quantum mechanincal energy prediction. We hypothesise that the reason for poor performance in this context is due to the large number of parameters and complex interactions which are more susceptible to overfitting on a biased dataset like \pdbbind.} as the base architecture for our 3D global models. We use the implementation in the \physicsml python package \anonref{physicsml}. As in previous work, we extract the protein pocket by selecting protein atoms within 5\AA\ of any ligand atom. The graph is constructed from the pocket and ligand atoms as nodes and a 5\AA\ cut-off is used to define the edges. We use one-hot encoded atomic numbers as node features and no edge features.

\paragraph{Pre-Trained EGNNs}
In this benchmark, we use a few different versions of the EGNN model. In addition to the basic model, we use two pre-trained versions. The first model, called \textbf{EGNN-QM}, is trained on ANI1ccx, a dataset of 500k small molecules with quantum mechanical energies computed at the ccsd(t) level. The knowledge learned from quantum mechanical interactions and internal energies should in principle better inform binding affinity prediction. The second model, \textbf{EGNN-DIFF}, is trained as a small molecule diffusion model on the QM9 dataset (as described in \cite{pmlr-v162-hoogeboom22a}). By pre-training on diffusing stable QM9 molecules, the model will have learned to distinguish between low and high energy conformations. In principle, this should allow it to better understand binding affinity interactions.

To transfer these models to the \pdbbind dataset, we use a two stage procedure. First, we freeze the backbone and add a new randomly initialised pooling head and train until convergence. Then, we unfreeze the backbone and train all parameters at a lower learning rate until convergence. 

Information about both pre-training strategies and transfer learning is available in the Supplementary Information.

\paragraph{Hydrogens} Previous works modelling binding affinity via GNNs have chosen to omit hydrogen atoms from the input graph (\cite{li2021SIGN,volkov2022frustration}). Since hydrogen atoms contribute significantly to binding via hydrogen bonds, we wanted to assess the effect of including hydrogen atoms as nodes in the graph. We benchmark the models with no hydrogen atoms (\textbf{None}), with only the polar hydrogen atoms (\textbf{Polar}), and with all hydrogens (\textbf{Explicit}).

\paragraph{\singlegraph vs. \multigraph} Finally, to probe whether the models learn to identify the interactions from the protein-ligand pose, we also train models on the pockets and ligands as separate graphs. We use the same backbone to generate embeddings for both and then combine these embeddings in a pooling head to make the final prediction. Practically, this removes any edges between protein and ligand nodes in the graph. We refer to these models, which treat proteins and ligands as separate graphs, as \textbf{\multigraph} models. On the other hand, the conventional models that operate on the interacting pose are referred to as \textbf{\singlegraph} models.

\subsubsection{\rfscore and \onion}
We include in our analysis two additional 3D global models, \rfscore and \onion. We select those models due to their performance on other splits of the \pdbbind dataset as indicated in the study by \cite{durant2023robustly}. Specifically, \rfscore was one of the top performer on the CASF-2016 split while the \onion model was superior on the 2019-Holdout and Peptides-Holdout sets relative to other models tested.

\subsubsection{Baseline models}
\paragraph{\ligandbias model}
To probe the ligand dataset bias in the benchmarking splits, we follow the work of \cite{durant2023robustly} and design a \ligandbias global model. 
This model is trained on the identical data used to train the global models (binding affinity measurements of ligands against different proteins), but only uses ligand-based features as input (analogous to \singleprotein models; see \ref{sec:methods:single_protein_local_models}). Effectively, this mixes binding affinity values of compounds measured against different proteins to probe the amount of dataset bias available in ligand information alone.

\paragraph{\mw model}
The molecular weight of a compound tends to be a strong predictor of its binding affinity, with larger compounds generally exhibiting stronger affinity \citep{olsson2008thermodynamics}. Within the context of drug discovery, it is particularly important to avoid building binding affinity models that strongly make use of the molecular weight property, as larger drug candidates have a higher probability of failure \citep{hopkins2014role}. In this study, we employ a \mw model that uses the molecular weight of the ligand as its sole input as a baseline. We use the same architectures and training data as for the \singleprotein local models.

\subsection{Splitting}

To build a robust benchmark and effectively probe model generalisation, we propose a new split of \pdbbind based on protein and ligand similarity, which we call the \textbf{\oursplit} split. Although many splits have been proposed, we believe that none achieve our goal of probing generalisation. Close inspection of the proposed splits \citep{volkov2022frustration,durant2023robustly,li2024leakproofpdbbindreorganized} shows non-negligible levels of similarity between train and test set, with all splits sharing some proteins (UniProts) across sets. Table \ref{table::overlap-table} shows the amount of UniProt overlap in previously proposed splits. Furthermore, the splits were not constructed to benchmark the variety of model families available (local vs. global, 3D vs. non-3D).

\begin{table}[h]
\centering  
\caption{Overview of UniProt overlap in previously proposed splits. The numbers denote the number of unique overlapping UniProts, overlapping test structures (out of total test structures), and overlapping train structures (out of total train structures).}
\begin{tabular}{lccc}  
\toprule  
\textbf{Split name} & \textbf{\# UniProts overlap} & \textbf{\# train overlap} & \textbf{\# test overlap} \\  
\midrule  
Post 2019 set \citep{volkov2022frustration} & 262 &  5520 / 16561 & 1004 / 1467 \\
CASF 2016 \citep{volkov2022frustration} & 67 &  3944 / 16561 & 281 / 282 \\
Zero-Ligand-Bias \citep{durant2023robustly} & 170 &  3724 / 17605 & 287 / 360 \\
LeakProof \pdbbind \citep{li2024leakproofpdbbindreorganized} & 172 &  1735 / 12923 & 3526 / 4751 \\
\oursplit (Ours) & 0 & 0 / 11022 & 0 / 1857 \\
\bottomrule  
\end{tabular}
\label{table::overlap-table}
\end{table}

\paragraph{\casestudy}
We design our newly proposed dataset splits such that we can compare between global and local models. This comparison is particularly relevant given that local \singleprotein models are still widely used in drug discovery projects, owing to their robustness, cost-effectiveness, and trainability on smaller datasets. To obtain a direct and fair comparison to these local models, we select eight proteins from the \pdbbind dataset which have more than 100 datapoints as our case study. The threshold of 100 points per protein is to allow enough data to train local models. In total, the \casestudy consists of 1857 structures. These proteins will be used to benchmark the generalisability of global models and also to train local models for each specific protein.

\begin{table}[h]  
\centering  
\caption{\casestudy}  
\begin{tabular}{lccc}  
\toprule  
\textbf{UniProt} & \textbf{HGNC} & \textbf{Number of structures} \\  
\midrule  
P00734 & F2           & 170 \\
P56817 & BACE1        & 343 \\
P24941 & CDK2         & 248 \\
O60885 & BRD4         & 199 \\
P00918 & CA2          & 425 \\
P07900 & HSP90AA1     & 172 \\
Q9H2K2 & TNKS2        & 113 \\
P00760 & N/A (Bovine) & 187 \\
\bottomrule  
\end{tabular}  
\end{table}

\begin{figure}
  \begin{minipage}[t]{.55\linewidth}
    \vspace{0pt}
    \includegraphics[width=\textwidth]{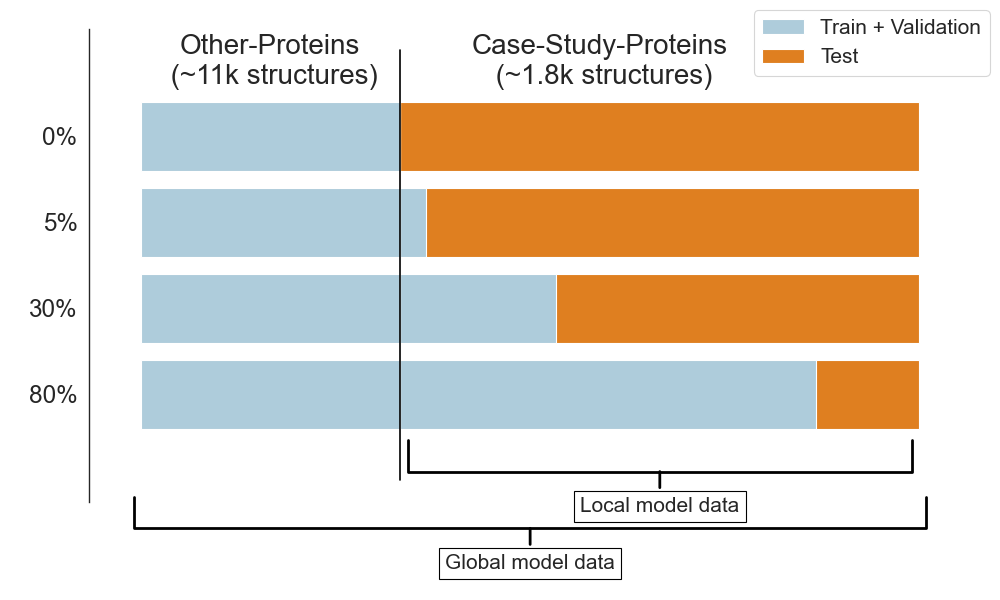} 
  \end{minipage}
  \begin{minipage}[t]{.4\linewidth}
    \vspace{25pt}
    \begin{tabular}{lcc}  
    \toprule  
    \textbf{Dataset} & \textbf{\# structures} \\  
    \midrule  
    \pdbbind & 19443 \\  
    \pdbbind prepared  & 18310 \\
    \cmidrule(lr){1-2}
    \oursplit & 12879\\
    - \casestudy & 1857 \\
    - \otherproteins  & 11022 \\
    \bottomrule  
    \end{tabular}  
  \end{minipage}
  \caption{Schematic of \oursplit benchmarking splits used in this study. Global models are trained on both train sets from the \casestudy and the \otherproteins.
    Local models are individually built for each of the eight proteins in the \casestudy split. They require a minimum set of already available ligands for a specific protein for training, thus can only be created for the 5\%, 30\%, 80\% splits and use only data from the \casestudy split.
    Note that bars are not to scale with number of samples.}
\label{fig::splits}
\end{figure}

\paragraph{Similarity filtering}
We apply two similarity filtering steps to the remaining structures in the \pdbbind dataset to create a subset that is dissimilar to the \casestudy set. We call this reduced dataset \otherproteins.

To account for protein similarity, we compute the similarity to the \casestudy using FoldSeek \citep{Van_Kempen2024-ql-foldseek}, which uses 3D structural and residue information to efficiently compute a similarity score [0, 1] between two protein structures. To probe the generalisability of the models, we filter out any structures which have more than 0.5 similarity to the \casestudy structures. Additionally, we compute the tanimoto similarity between the \casestudy ligands and the ligands of the remaining structures. We filter out any structures with ligands with more than 0.5 tanimoto similarity. A total of 5431 similar structures are removed and leaves 11022 structures in the \otherproteins set.

\paragraph{\oursplit 0\%, 5\%, 30\%, 80\%}
We aim to examine the change in model performance as the amount of data increases, from low data regimes (when binding affinity values are available for approximately 0 to 30 ligands for a specific protein) in comparison to a medium data scenario where increasingly more data is available.

We use the \oursplit split (1875 \casestudy structures and 11022 \otherproteins structures) to construct the benchmarking splits as follows. We stratify the \casestudy data by protein and split the structures by tanimoto ligand similarity\footnote{We note that tanimoto splits are harder than scaffold splits since scaffold which technically are distinct can still have high tanimoto similarity.} with increasing fraction of training data, 5\%, 30\%, 80\%. At each percentage, we generate three folds with a different starting seed ligand for the similarity splits. Local models (\singleprotein and \mw) are trained on these splits for each protein individually. For the global models (\rfscore, \onion, EGNN, \ligandbias baseline), we further augment the train sets of these splits with the \otherproteins. Additionally, we construct the 0\% split where all \casestudy are in the test set and only the \otherproteins are in the train set. This is to probe generalisation to completely new proteins. The final benchmarking splits are shown schematically in Figure~\ref{fig::splits}. In absolute numbers, those splits correspond to the following number of training samples per protein: 11$\pm$5 (5\%), 69$\pm$30 (30\%), 185$\pm$83 (80\%).

\subsubsection{Training and validation}

For each split (0\%, 5\%, 30\%, and 80\%), we perform a cross-validation split of (random 80:20 split) for model selection. The models are then retrained on the combined train and validation set and the performance is measured on the test set. 

For deep learning models (EGNN architectures), we additionally use a random 80:20 split of the train set for early stopping.

\subsubsection{Metrics}

Typically, binding affinity models are assessed using the Pearson correlation coefficient, which measures the correlation between predicted and actual binding affinity values. However, this doesn't measure the absolute predictive performance, which is crucial in real-world drug discovery for optimising multi-parameter objectives. Therefore, here we focus on the absolute $R^2$ metric for benchmarking performance, while also providing root mean squared error and Pearson correlation results in the Supplementary Information.

On the \oursplit test sets, we compute the metrics in two different ways. The \textbf{overall} performance refers to metrics computed on the predicted and actual binding affinity values across all eight proteins. Mixing predictions of all the different protein-ligand pairs is the common approach when reporting results on the \pdbbind dataset \citep{meli2022scoring}. We also report the performance \textbf{stratified by protein}, where metrics are calculated individually for each protein.

%% file: sections/3_results.tex
\section{Results}

We now present the results of the benchmarking. First, we present the results and comparisons across model families. We follow this with more detailed analyses of the 3D EGNN models.

\subsection{Model family comparisons}

\begin{figure} 
  \centering
  \includegraphics[width=\textwidth]{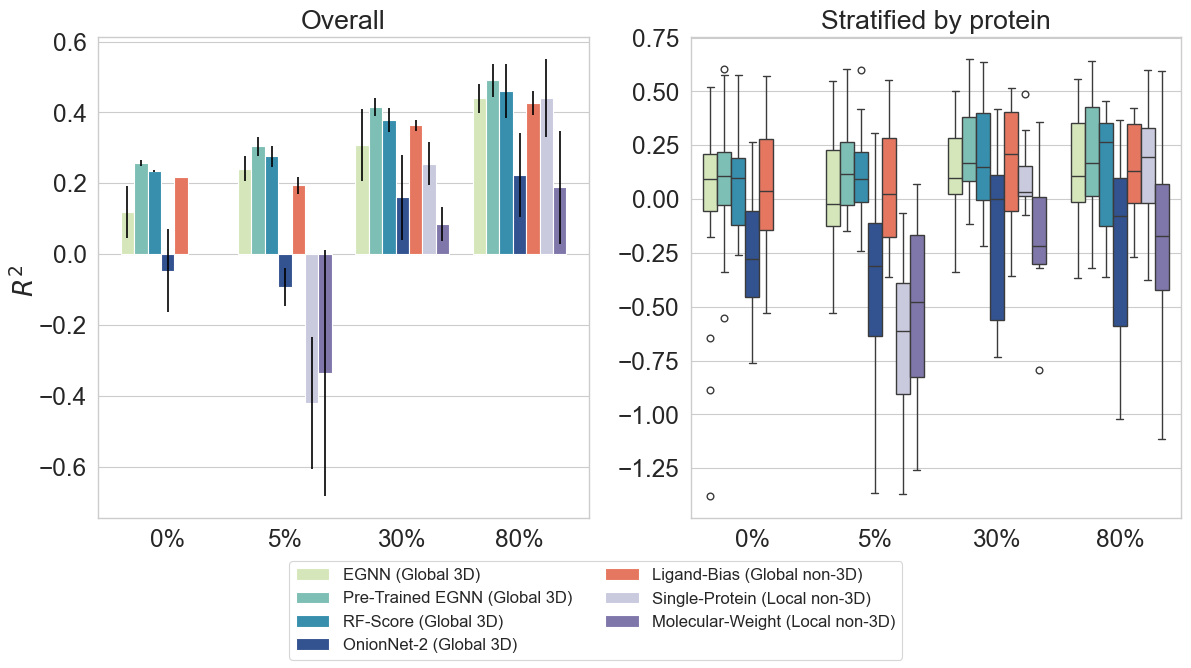} 
  \caption{Overall and stratified performance at increasing train data fraction for different model families. In the low data regime, global 3D models outperform local models. Left: The error bars denote the standard deviation across the three test folds. Right: The boxplots represent the performance distribution over the eight proteins in the \casestudy set.}
  \label{fig:overall_metrics}
\end{figure}

The results reveal a clear advantage for global 3D models over local models in low data regimes. As seen in Figure \ref{fig:overall_metrics}, the global 3D models (EGNN, pre-trained EGNN, and \rfscore) have moderate generalisation even with 0\% protein-specific training data and outperform the \singleprotein model at lower train data fractions (5\% and 30\%). With enough data (80\%), most models plateau at similar performance (EGNN, pre-trained EGNN, \rfscore, \singleprotein). We hypothesise that this is due to the difficulty of the tanimoto similarity splits. At low data levels, the local models not only have very little data to learn from but also a train set which is very dissimilar to the test ligands. In drug discovery settings where generalisation is important, this benchmark clearly highlights the importance of global models.

Notably, the pre-trained EGNN models outperform all other models and baselines when considering the overall performance (Figure \ref{fig:overall_metrics}, left). See \secref{subsec::EGNN_additions} for a detailed analysis of their performance.

One thing to note is the relatively good performance of the \ligandbias model (red bar in Figure~\ref{fig:overall_metrics}). Although not as good as the global 3D models, the \ligandbias model is able to generalise to the unseen proteins. We hypothesise that this is due to the model picking up on functional group importance in the ligands, given that a lot of binders rely on generic functional groups.

\begin{figure} 
  \centering
  \includegraphics[width=\textwidth]{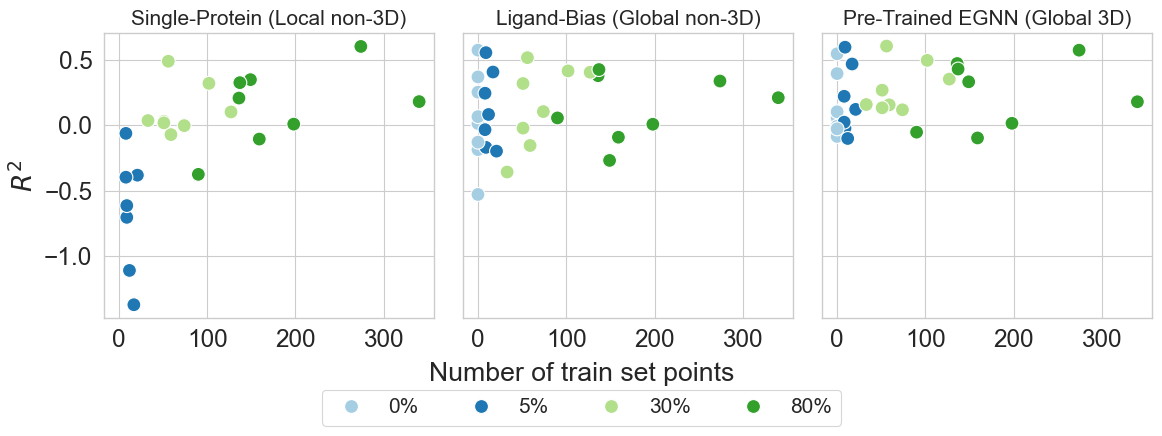} 
  \caption{Effect of number of training data points on performance. Each point represents a protein from the eight case-study proteins. The global models show a clear advantage at low data regimes.}
  \label{fig::performance-scatter}
\end{figure}

It is also interesting to note that, as seen in Figure~\ref{fig::performance-scatter}, the performance of the global 3D models is relatively consistent with the level of training data (improving slightly) whereas the \singleprotein local model suffers greatly at low data, dramatically improving with increasing amounts of training data.

\begin{figure}[!htb]
  \centering
  \includegraphics[width=\textwidth]{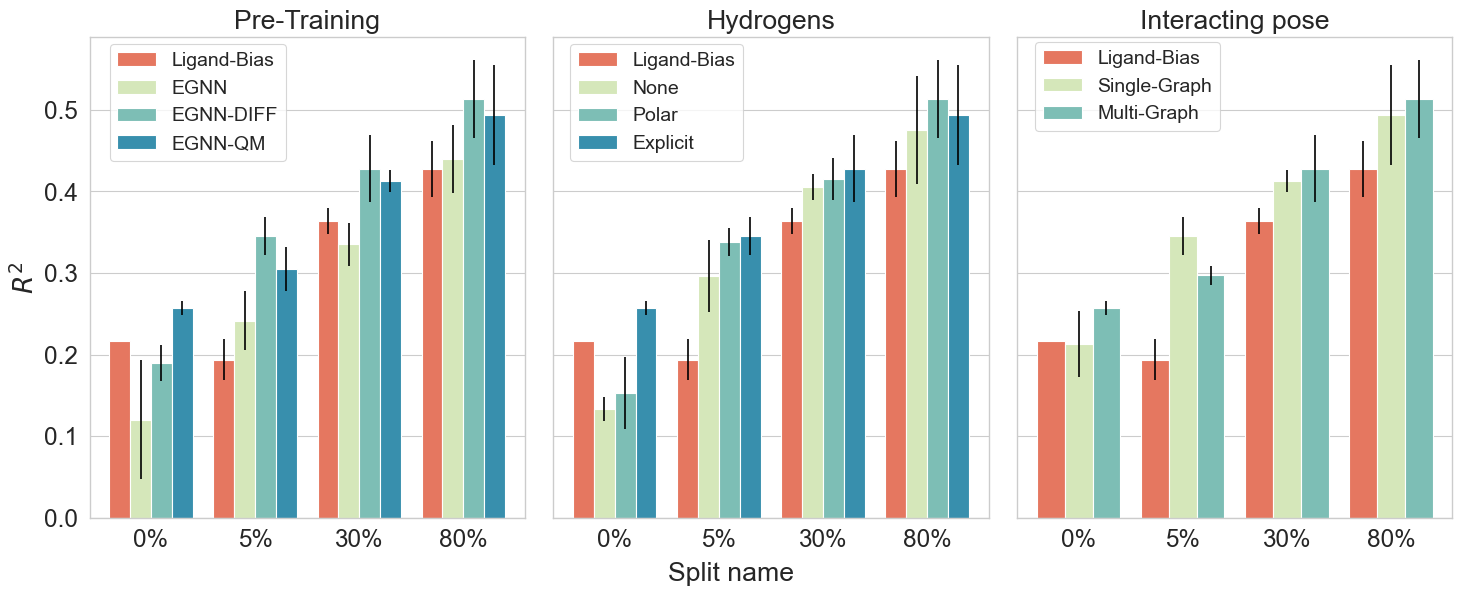} 
  \caption{Effect of the EGNN additions proposed in this study on model performance. The overall performance across all eight proteins in the \casestudy set is reported. The error bars denote the standard deviation across the three test folds. In the 0\% split case, there is a only a single test fold. Due to the non-deterministic nature of training, variation in performance is due to training the same EGNN model three times. \textbf{Pre-training}: Quantum mechanical pre-training provides the greatest advantage, followed closely by diffusion pre-training.
  \textbf{Hydrogens}: Including explicit hydrogens is very important at low data levels.
  \textbf{Interacting pose}: No consistent pattern when comparing single-graph versus multi-graph. }
  \label{fig::gnn-benchmarks}
\end{figure}

\subsection{Improvements to EGNNs}
\label{subsec::EGNN_additions}

We now present a detailed analysis of the global 3D EGNN models.

\subsubsection{Pre-Training}

First, we look at the effect of pre-training on model performance. As Figure \ref{fig::gnn-benchmarks} shows, pre-training significantly improves model performance. Quantum mechanical pre-training provides the largest advantage followed closely by diffusion pre-training. As expected, the advantages are more noticeable at low levels of training data, gradually fading out as more training data is added. To our knowledge, this is the first application of using pre-trained models for 3D binding affinity prediction.

\subsubsection{Hydrogens}

Next, we assess the effect of including hydrogen atoms on model performance. To our knowledge, in all previously published work, hydrogen atoms were ignored in the structures. It was not clear if this occurred due to better model performance on the benchmarks or due to knowledge of the poorly prepared \pdbbind structures. As described in section \ref{dataset preparation}, we noticed that the structures in \pdbbind did not have consistent hydrogen preparation and used CCDC software to add them consistently.

As can be seen in Figure \ref{fig::gnn-benchmarks}, hydrogens are very import at low data levels for generalisation. With more data, their effect becomes negligible. This is very important to keep in-mind for building models that generalise to new proteins and novel ligand chemical space.

\subsubsection{\singlegraph vs. \multigraph}

Finally, we look at the effect of the interacting pose on performance.
In the default architecture (single-graph), the ligand and pocket are given to the model as a single 3D graph. In the multi-graph architecture, the ligand and the protein are first encoded with separate graphs. In theory, we expect the single-graph to outperform the multi-graph architecture. 

As we can see in Figure \ref{fig::gnn-benchmarks}, using the pose information (single-graph model) does not necessarily improve the model performance. This could either be due to the model being able to infer the interactions without the exact pose or not properly learning the interactions in the first place (and so we do not notice its effect). Either way, since these models outperform the \ligandbias model, they must be using the protein information in some way. Further investigation is required to understand what the models are learning. 

%% file: sections/4_discussion.tex
\section{Discussion}

In this paper, we demonstrate that for binding affinity prediction on new proteins and chemical spaces, global 3D models outperform local models on the \pdbbind dataset. Furthermore, we show that explicit hydrogen atoms in the structures and novel pre-training strategies using quantum mechanical data and diffusion modelling provide performance improvements in low data regimes for GNNs.

There are limitations to the above benchmark. This work focuses on the \pdbbind dataset, a dataset comprising only crystal structures. We explicitly chose to do so to eliminate any sources of noise or error from computationally generated structures and poses. However, this has two drawbacks. First, the structures and chemical space of ligands in the dataset are not representative of the distribution of ligands in a real-world drug discovery projects. The relatively good performance of the \ligandbias model indicates the lack of diversity in the ligand distribution. Second, since crystal structures can only be obtained for binding ligands, this benchmark does not probe the performance of models for non-binding ligands. This is important for virtual high throughput screens where many ligands might not be binders.

In light of these limitations, it is crucial to replicate a similar analysis on additional datasets, ideally resembling real-world drug discovery datasets. If insights from this study are confirmed with more datasets, this research could represent a significant stride towards developing more universally applicable 3D binding affinity models that leverage pre-training strategies.


%% file: sections/5_acknowledgements.tex
\makeatletter  
\if@submission
\else
\section*{Acknowledgements}
This work benefited greatly from the input and feedback of many colleagues. We are grateful to Ben Butt, Gail Bartlett, Richard Bradshaw, Douglas Pires, David Errington, Daniel Cutting, Emil Nichita, Jonathan Harrison, Constantin Schneider, Andrew Wedlake, Jody Barbeau for useful discussions.
\fi  
\makeatother

%% file: sections/supplementary_info.tex
\begin{center}
\textbf{\large Supplemental Information}
\end{center}

\section{\singleprotein hyperparameters}

We used classical architectures and features for the \singleprotein models. All the models and features below can be found in the python package \molflux (\anonref{molflux}).

\begin{table}[h]  
\centering  
\caption{\singleprotein models}  
\begin{tabular}{lccc}  
\toprule  
\textbf{Model Name} & \textbf{hyperparameters}\\  
\midrule  
Random Forest & $\mathtt{n\_ estimators}=500$ \\
\midrule  
XGBoost & \makecell{$\mathtt{learning\_ rate}=0.2$ \\ $\mathtt{subsample}=1$ \\ $\mathtt{max\_ depth}=6$} \\
\midrule  
CatBoost & $\mathtt{random\_ state}=0$ \\
\midrule  
Support Vector Regressor & $\mathtt{kernel}=\mathtt{rbf}$ \\
\bottomrule  
\end{tabular}  
\end{table}

\begin{table}[h]  
\centering  
\caption{\singleprotein features}  
\begin{tabular}{lccc}  
\toprule  
\textbf{Feature name} & \textbf{Comments}\\  
\midrule  
Molecular Weight &  \\
\midrule  
Molecular descriptors (MD) & \makecell{xLogP, aromatic ring count \\ molecular weight, num acceptors \\ num donors, rotatable bonds, tpsa \\  \\ All descriptors were normalised  \\ by the train set mean and variance}\\
\midrule  
ECFP & circular 2048 fingerprints \\
\midrule  
FCFP & \\
\midrule  
Topological torsion & \\
\midrule  
ECFP + MD &  \\
\midrule  
FCFP + MD &  \\
\midrule  
Topological torsion + MD & \\
\bottomrule  
\end{tabular}  
\end{table}

\section{EGNN models}

We used the following hyperparameters for the EGNN models.

\begin{table}[h]  
\centering  
\caption{\singleprotein models}  
\begin{tabular}{lccc}  
\toprule  
\textbf{Hyperparameter} & Value \\  
\midrule  
$\text{num}\_\text{layers}$ & 5 \\
$\text{c}\_\text{hidden}$ & 128 \\
$\text{num}\_\text{rbf}$ & 8 \\
$\text{pool}\_\text{type}$ & sum \\
Activation & SiLU \\
Loss & MSELoss \\
Optimizer & AdamW, LR$=5\times10^{-4}$ \\
Scheduler & ReduceLROnPlateau \\
\bottomrule  
\end{tabular}  
\end{table}

\section{Pre-training EGNN models}

\subsection{Quantum mechanical energy}

The EGNN-QM model was pre-trained on the ccsd(t) energy of the ANI1ccx dataset. This is a dataset of roughly 500k small molecules. As is common in the domain of neural network potentials, the model was trained to predict the interaction energy (the total energy minus the self atomic energies of the atoms). The final root mean squared error of the model was 4kcal/mol. 

\subsection{Small molecule diffusion}

The EGNN-DIFF model was trained to generate small molecules from the QM9 dataset (small molecule with up to 9 heavy atoms). This was carried out as described in the original work by \cite{pmlr-v162-hoogeboom22a}. 

For use in a predictive setting, the coordinate updates in the diffusion model were turned off and a predictive pooling head was added. 

\subsection{Transfer learning}

We match the backbones of the pre-trained models to freshly initialized models (the pooling heads are not matched, they remain randomly initialized). The backbone is then trained until convergence using early stopping at a learning rate or $5\times10^{-4}$. The backbone is then unfrozen and the entire model is fine-tuned at a lower learning rate $1\times10^{-4}$.

\subsection{Training resources}

Each EGNN model training takes $\sim$6 hours on a single A10 GPU.

\section{Metrics}

Below are the explicit equations for the metrics used in the benchmarks

\begin{equation*}
    \textnormal{Pearson Correlation Coefficient} = \frac{\sum_{i} (x_i - \bar{x})(y_i - \bar{y})}{\sqrt{\sum_{i} (x_i - \bar{x})^2} \sqrt{\sum_{j} (y_j - \bar{y})^2}}
\end{equation*}

\begin{equation*}
    R^2 = 1 - \frac{\sum_{i} (x_i - y_i)^2}{\sum_{j} (y_j - \bar{y})^2}
\end{equation*}

\section{Additional plots}

\begin{figure}[!htb]
  \centering
  \includegraphics[width=\textwidth]{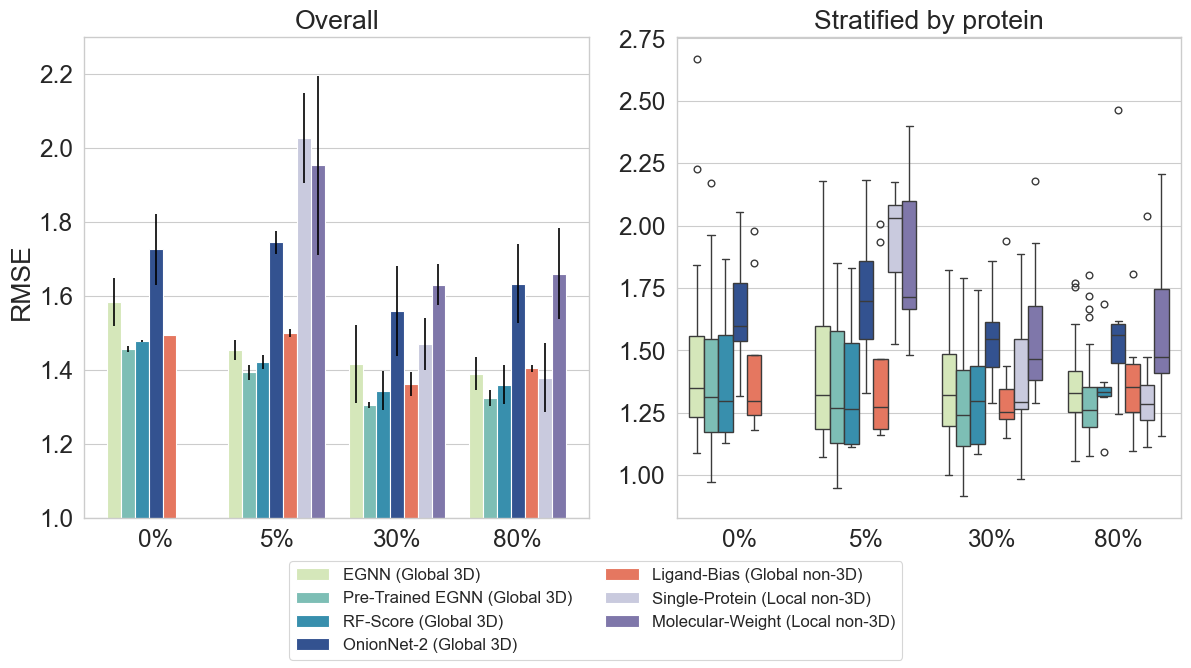} 
  \caption{Overall and stratified performance at increasing train data fraction for different model families. In the low data regime, global 3D models outperform local models. Left: The error bars denote the standard deviation across the 3 test folds. Right: The boxplots represent the performance distribution over the eight proteins in the \casestudy set.}
\end{figure}

\begin{figure}[!htb]
  \centering
  \includegraphics[width=\textwidth]{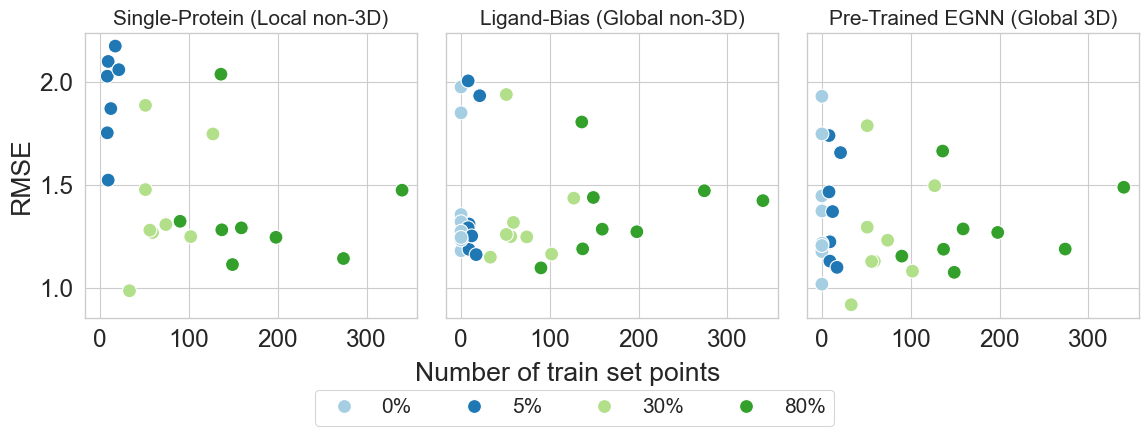} 
  \caption{Effect of number of training data points on performance. Each point represents a protein from the 8 case-study proteins. The global models show a clear advantage at low data regimes.}
\end{figure}

\begin{figure}[!htb]
  \centering
  \includegraphics[width=\textwidth]{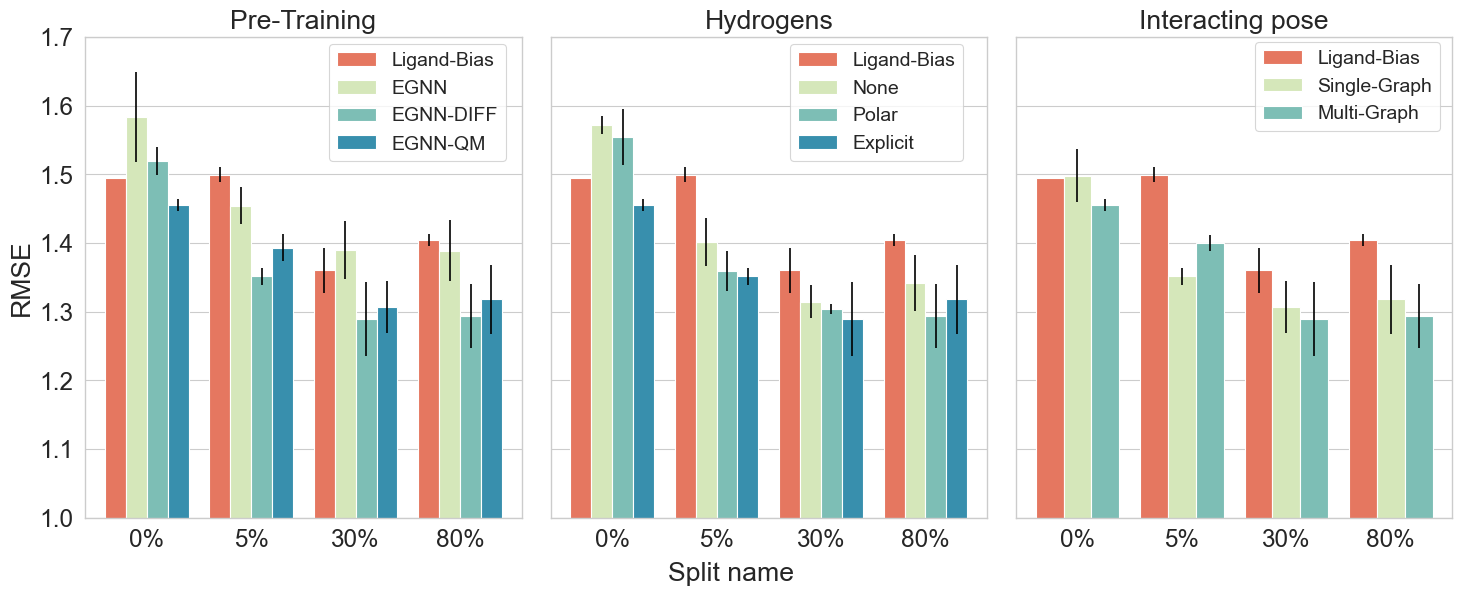} 
    \caption{Effect of the EGNN additions proposed in this study on model performance. The overall performance across all eight proteins in the \casestudy set is reported. The error bars denote the standard deviation across the 3 test folds. \textbf{Pre-training}: Quantum mechanical pre-training provides the greatest advantage, followed closely by diffusion pre-training.
  \textbf{Hydrogens}: Including explicit hydrogens is very important at low data levels.
  \textbf{Interacting pose}: No consistent pattern becomes apparent when comparing single-graph versus multi-graph. }
\end{figure}

\begin{figure}[!htb]
  \centering
  \includegraphics[width=\textwidth]{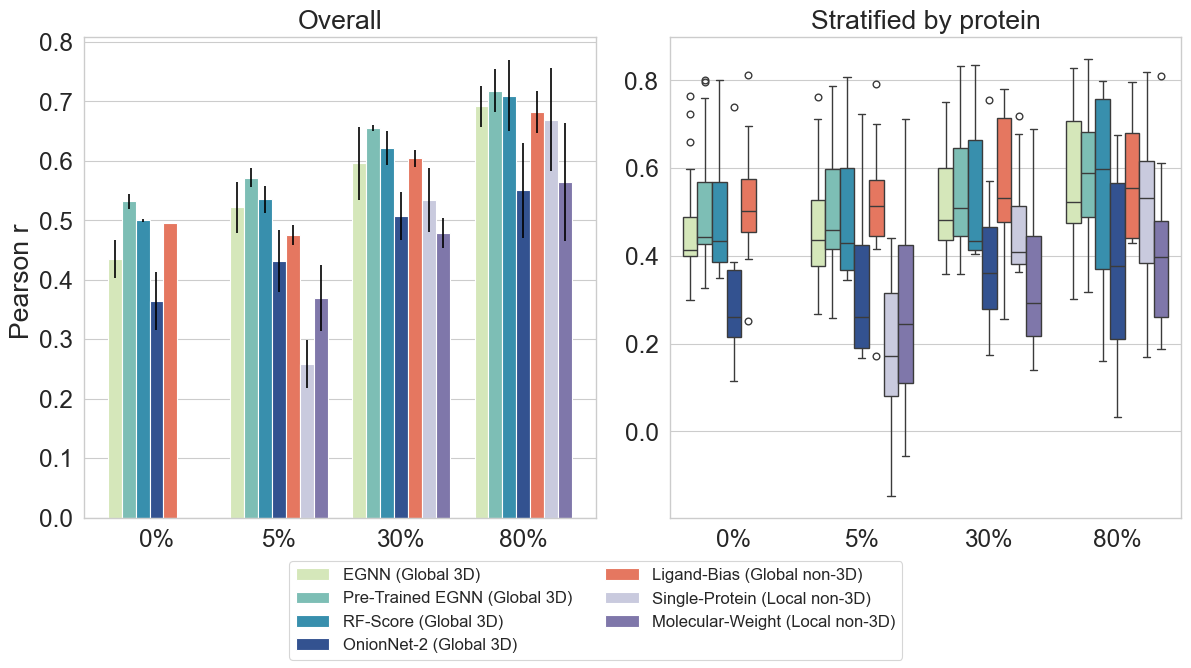} 
  \caption{Overall and stratified performance at increasing train data fraction for different model families. In the low data regime, global 3D models outperform local models. Left: The error bars denote the standard deviation across the 3 test folds. Right: The boxplots represent the performance distribution over the eight proteins in the \casestudy set.}
\end{figure}

\begin{figure}[!htb]
  \centering
  \includegraphics[width=\textwidth]{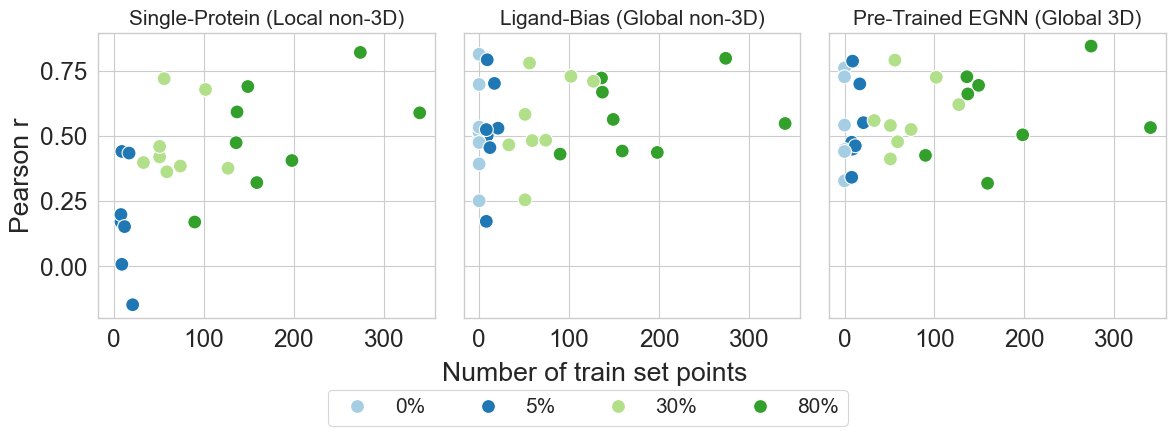} 
  \caption{Effect of number of training data points on performance. Each point represents a protein from the 8 case-study proteins. The global models show a clear advantage at low data regimes.}
\end{figure}

\begin{figure}[!htb]
  \centering
  \includegraphics[width=\textwidth]{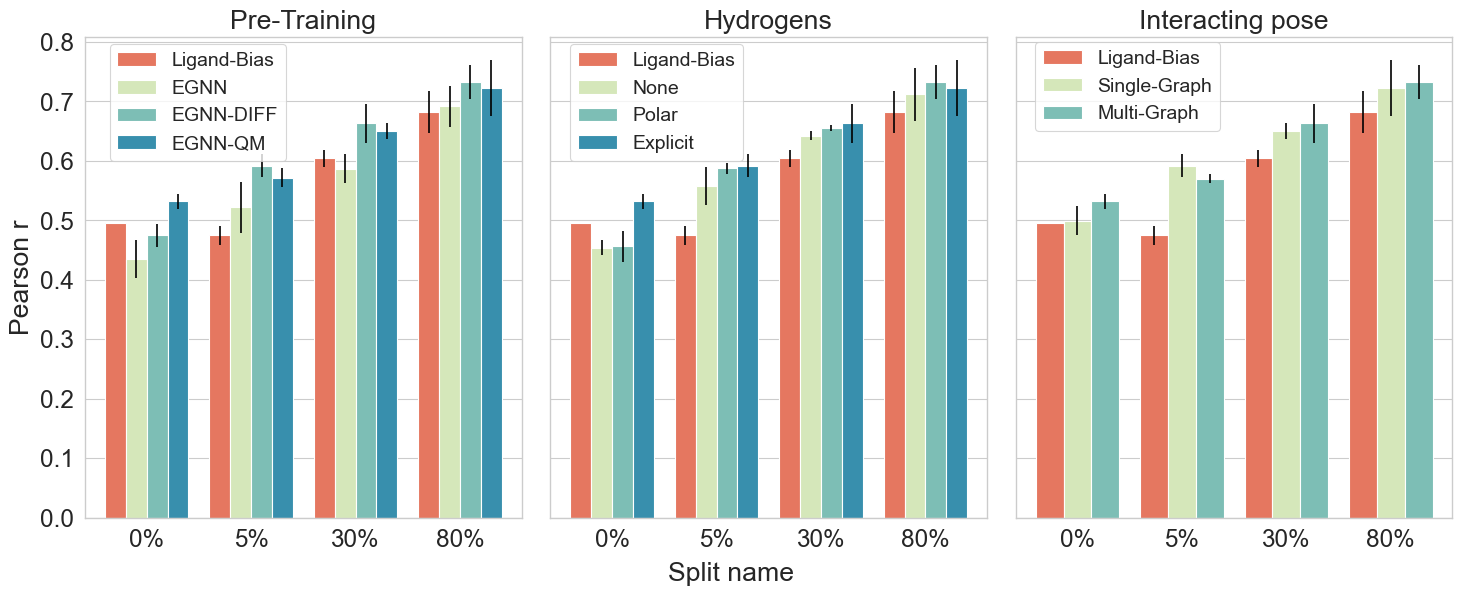} 
    \caption{Effect of the EGNN additions proposed in this study on model performance. The overall performance across all eight proteins in the \casestudy set is reported. The error bars denote the standard deviation across the 3 test folds. \textbf{Pre-training}: Quantum mechanical pre-training provides the greatest advantage, followed closely by diffusion pre-training.
  \textbf{Hydrogens}: Including explicit hydrogens is very important at low data levels.
  \textbf{Interacting pose}: No consistent pattern becomes apparent when comparing single-graph versus multi-graph. }
\end{figure}